\documentclass[12pt]{article}

\usepackage{arxiv}

\usepackage[utf8]{inputenc} 
\usepackage[T1]{fontenc}    
\usepackage{hyperref}
\hypersetup{colorlinks,allcolors=green}
\usepackage{url}            
\usepackage{booktabs}       
\usepackage{amsfonts}       
\usepackage{nicefrac}       
\usepackage{microtype}      
\usepackage{lipsum}
\usepackage{graphicx}
\usepackage{setspace} 
\usepackage{comment}
\usepackage{float}
\usepackage{amsmath,amssymb,amsfonts}

\graphicspath{ {./images/} }

\title{Enhancing Facial Classification and Recognition using 3D Facial Models and Deep Learning}

\author{
 Houting Li\thanks{These authors contributed equally to this work.} \\
  Department of Mathematics\\
  The Chinese University of Hong Kong\\
  Hong Kong \\
  \texttt{hooting@link.cuhk.edu.hk} \\
   \And
 Mengxuan Dong\footnotemark[1] \\
  Department of Mathematics\\
  The Chinese University of Hong Kong\\
  Hong Kong \\
  \texttt{dongmengxuan@link.cuhk.edu.hk} \\
  \And
 Lok Ming Lui \\
  Department of Mathematics\\
  The Chinese University of Hong Kong\\
  Hong Kong \\
  \texttt{lmlui@math.cuhk.edu.hk} \\
}

\date{}

\microtypesetup{nopatch=footnote}
\begin{document}
\onehalfspacing
\maketitle
\begin{abstract}
Facial analysis and classification are critical tasks with diverse applications, including security, healthcare, and human-computer interaction. Traditional approaches often rely on 2D information extracted from facial images, which may fail to capture the geometric intricacies of human faces, leading to suboptimal classification accuracy. To address this limitation, we propose a novel approach to tackle various classification tasks by integrating the power of the 3D Morphable Model (3DMM) with the versatility of the ResNet architecture. The 3DMM serves as a pivotal tool for extracting shape and expression information from 2D facial images, representing a facial surface as a linear combination of shape and texture bases derived from a statistical analysis of a 3D facial database. Additionally, an approach utilizing information from images captured at multiple angles enhances the accuracy of 3DMM feature extraction. ResNet operates as our classification model, leveraging these features to systematically extract the most salient information suited for distinct classification tasks. Our method achieves significant improvements in classification accuracy, outperforming state-of-the-art methods. Specifically, we achieve 100\% accuracy in individual classification, 95.4\% in gender classification, and 83.5\% in expression classification. These results underscore the effectiveness of combining 3D facial insights with deep learning techniques and highlight the potential of our approach to advance the field of facial recognition and analysis.
\end{abstract}

\newpage

\section{Introduction}

Facial analysis and classification have become essential areas of research in the field of computer vision due to their wide-ranging applications in security, healthcare, entertainment, and human-computer interaction. The ability to accurately classify and recognize facial features enables numerous practical applications, such as personalized customer experiences, accurate emotion recognition, enhanced human-computer interaction, and secure authentication systems. For example, facial recognition systems are widely deployed in surveillance for security purposes, while emotion recognition has gained traction in healthcare for monitoring mental health conditions. Despite these advancements, achieving robust and accurate facial classification remains a challenging task due to the inherent variability in human facial geometry and the influence of external factors such as lighting conditions, pose variations, occlusions, and facial expressions.

Traditional methods for facial classification and recognition typically rely on 2D image-based approaches, which extract features directly from pixel values or preprocessed image representations. Convolutional Neural Networks (CNNs) have proven to be the cornerstone of these approaches, with well-established models such as FaceNet~\cite{schroff2015facenet} and ResNets~\cite{he2016deep} demonstrating exceptional performance in facial analysis tasks. These models learn hierarchical feature representations, enabling them to capture complex patterns in 2D images. However, the reliance on 2D data introduces inherent limitations. Since 2D images are projections of 3D objects, they fail to fully capture the geometric and structural details of the human face. Consequently, critical spatial and geometric information, such as subtle variations in facial contours, may be lost. This loss of information can lead to suboptimal performance in tasks such as individual identification, emotion recognition, and gender classification, particularly under challenging conditions like extreme poses or occlusions.

To address these limitations, the use of 3D facial information has gained significant attention in recent years. Unlike 2D data, 3D facial models inherently capture the geometric and structural details of human faces, making them more robust to variations in pose, lighting, and expression. Among the various techniques for modeling 3D faces, the 3D Morphable Model (3DMM) has emerged as a powerful generative method. By statistically analyzing a database of 3D facial scans, the 3DMM constructs shape and texture bases that can represent any 3D facial surface as a linear combination of these bases~\cite{blanz1999morphable}. This capability allows for the reconstruction of 3D facial geometry from 2D images, effectively bridging the gap between 2D visual data and 3D geometric representations. Additionally, the 3DMM provides a compact and discriminative representation of facial geometry through its shape coefficients, which can be directly used as input features for classification tasks.

In this work, we propose a novel approach to enhance facial classification and recognition tasks by combining the power of the 3DMM with the versatility and robustness of deep learning models. Specifically, we leverage the 3DMM to extract shape and expression information from 2D facial images, providing a rich set of geometric features that serve as input to a ResNet-based classification model. Unlike many existing methods that analyze geometric properties directly from reconstructed 3D surfaces, our approach focuses on the shape coefficients associated with the 3DMM. These coefficients encapsulate essential geometric details of the face in a compact and discriminative form, making them highly suitable for classification tasks. By systematically extracting and utilizing the most informative features, our approach improves the performance of classification models during both training and validation stages.

A key innovation of our method is the incorporation of images captured from multiple angles for facial classification to improve the accuracy of 3DMM feature extraction, using the method proposed in \cite{deng2019accurate}. By leveraging information from diverse viewpoints, we mitigate the ambiguities that arise from single-view 2D images, achieving a more robust estimation of 3D facial geometry. This, in turn, enhances the quality of the extracted shape coefficients and boosts the performance of the classification model. Our method systematically selects the most salient features for each classification task, ensuring that the extracted information is optimally tailored to the specific requirements of individual, gender, and expression classification.

The integration of 3D facial insights with deep learning models represents a significant advancement in the field of facial analysis. Deep learning, particularly convolutional neural networks like ResNet~\cite{he2016deep}, has revolutionized computer vision by enabling the automated learning of hierarchical feature representations from raw data. By combining the discriminative power of ResNet with the geometric richness of 3DMM, our approach achieves notable improvements in classification accuracy. Specifically, we demonstrate state-of-the-art performance across multiple tasks, achieving 100\% accuracy in individual classification, 95.4\% in gender classification, and 83.5\% in expression classification.

The key contributions of this work can be summarized as follows:
\begin{itemize}
    \item We propose a framework that combines 3DMM shape coefficients extracted from 2D images taken from multiple views with a deep learning-based classification model to perform facial analysis tasks such as individual recognition, gender classification, and expression classification. By leveraging multi-view images, we enhance the accuracy of 3DMM feature extraction, improving the robustness and reliability of the method.
    \item We systematically analyze the effectiveness of the extracted features for distinct classification tasks, demonstrating their suitability and impact on improving classification accuracy across multiple benchmarks.
\end{itemize}

The remainder of this paper is organized as follows. Section~\ref{sec:relatedwork} reviews related work in the domains of facial classification and 3D facial modeling. Section~\ref{sec:methodology} describes the methodology, including the 3DMM formulation, feature extraction process, and ResNet-based classification model. Section~\ref{sec:experiments} presents the experimental setup and results, highlighting the performance of our approach compared to state-of-the-art methods. Finally, Section~\ref{sec:conclusion} concludes the paper and discusses potential directions for future research.



\section{Previous Work}
\label{sec:relatedwork}
Advancements in facial classification and recognition have been significantly influenced by deep learning approaches. Yi Sun, Xiaogang Wang, and Xiaoou Tang, in their work DeepID\cite{sun2014deep} introduced Deep hidden IDentity features. These features, derived from deep convolutional networks, are learned through multi-class face identification tasks and can generalize to tasks like verification and identities unseen in the training set. With the capability to recognize around 10,000 face identities, these features achieve high verification accuracy, even with weakly aligned faces.

Another notable contribution is from Y Taigman, M Yang, MA Ranzato, and L Wolf in their study Deepface\cite{taigman2014deepface} They revisited the alignment and representation stages of face recognition by employing explicit 3D face modeling and a nine-layer deep neural network. This approach, trained on a large dataset of four million facial images, significantly reduced the error in face verification, approaching near-human-level performance. This is a cornerstone work to incoporate 3D information in human facial classifications.

Following these groundbreaking studies, other research has focused on more dataset-specific advancements. Eng et al.\cite{eng2019facial} used histogram of oriented gradients (HOG) and support vector machines (SVM) for facial expression recognition on the KDEF dataset, achieving an accuracy of 80.95\%. In gender classification, Thomaz and Giraldi\cite{thomaz2010new} implemented a novel ranking method for principal component analysis (PCA) on the FEI database, achieving around 97\% accuracy.


\section{Proposed Method}
\label{sec:methodology}
The parameters $\alpha$ and $\beta$ used for classification are derived through 3D Facial Model Reconstruction. In Section 5.1, we adopt R-Net and C-Net, as proposed in \cite{deng2019accurate}, to extract $\alpha$ and $\beta$, where $\alpha$ primarily influences shape and $\beta$ affects expression. R-Net is employed to extract parameters essential for 3D facial reconstruction, while C-Net aggregates information from multiple views. The extracted parameters $\alpha$ and $\beta$ are then used to train ResNet-34 for different classification tasks. We now provide a detailed description of each component.

\begin{figure}[h]\label{individualgender}
  \centering
  \includegraphics[width=\textwidth]{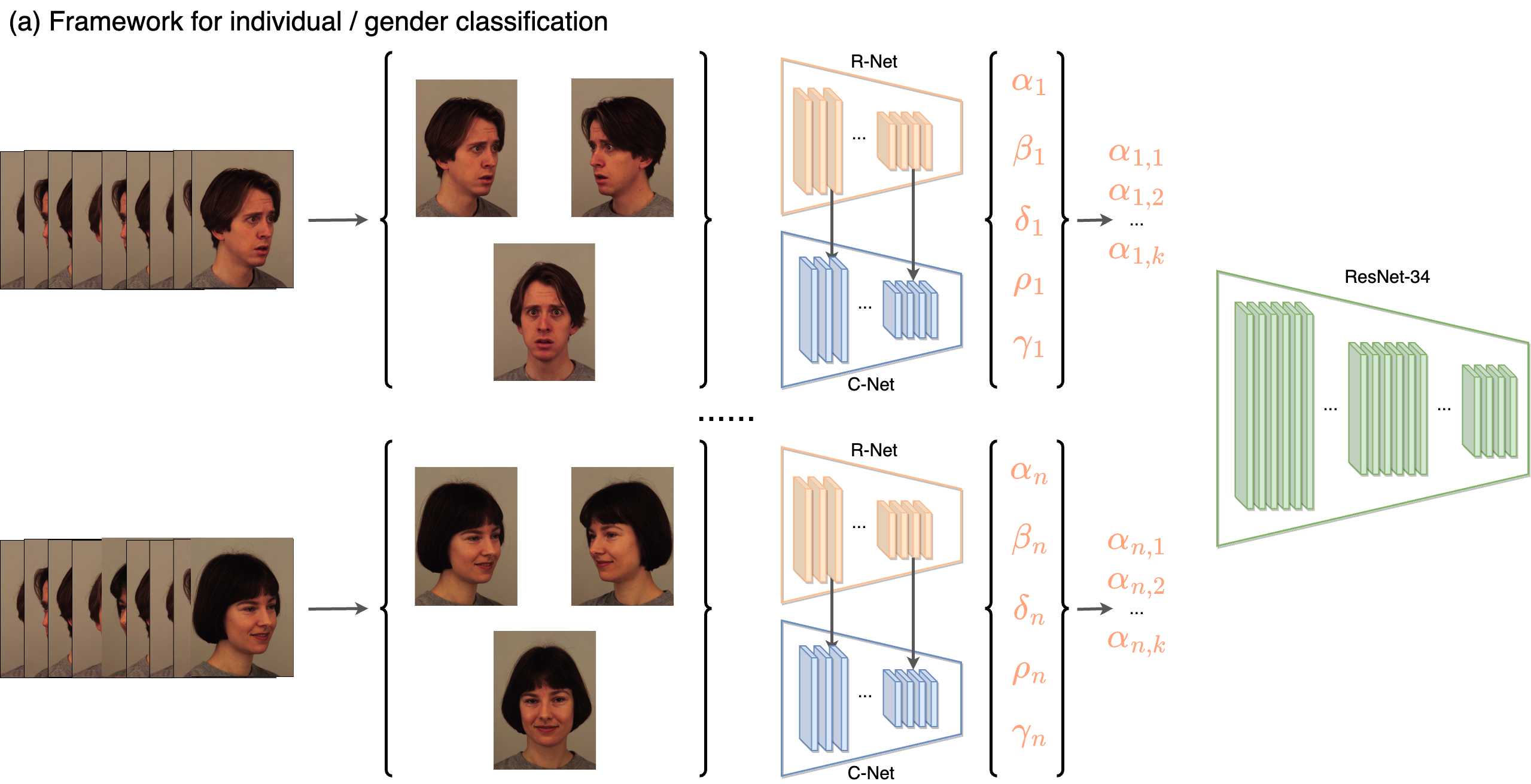}
  \caption{Framework for Individual/Gender Classification using Convolutional Neural Networks. The process starts with multiple angle images of individuals, which are fed into the R-Net to extract feature vectors. C-Net is used to make an accurate reconstruction from multiple images. After generated $\alpha$, it will be used to train ResNet-34 for classification}
\end{figure}
\newpage

\begin{figure}[h]\label{expression}
  \centering
  \includegraphics[width=\textwidth]{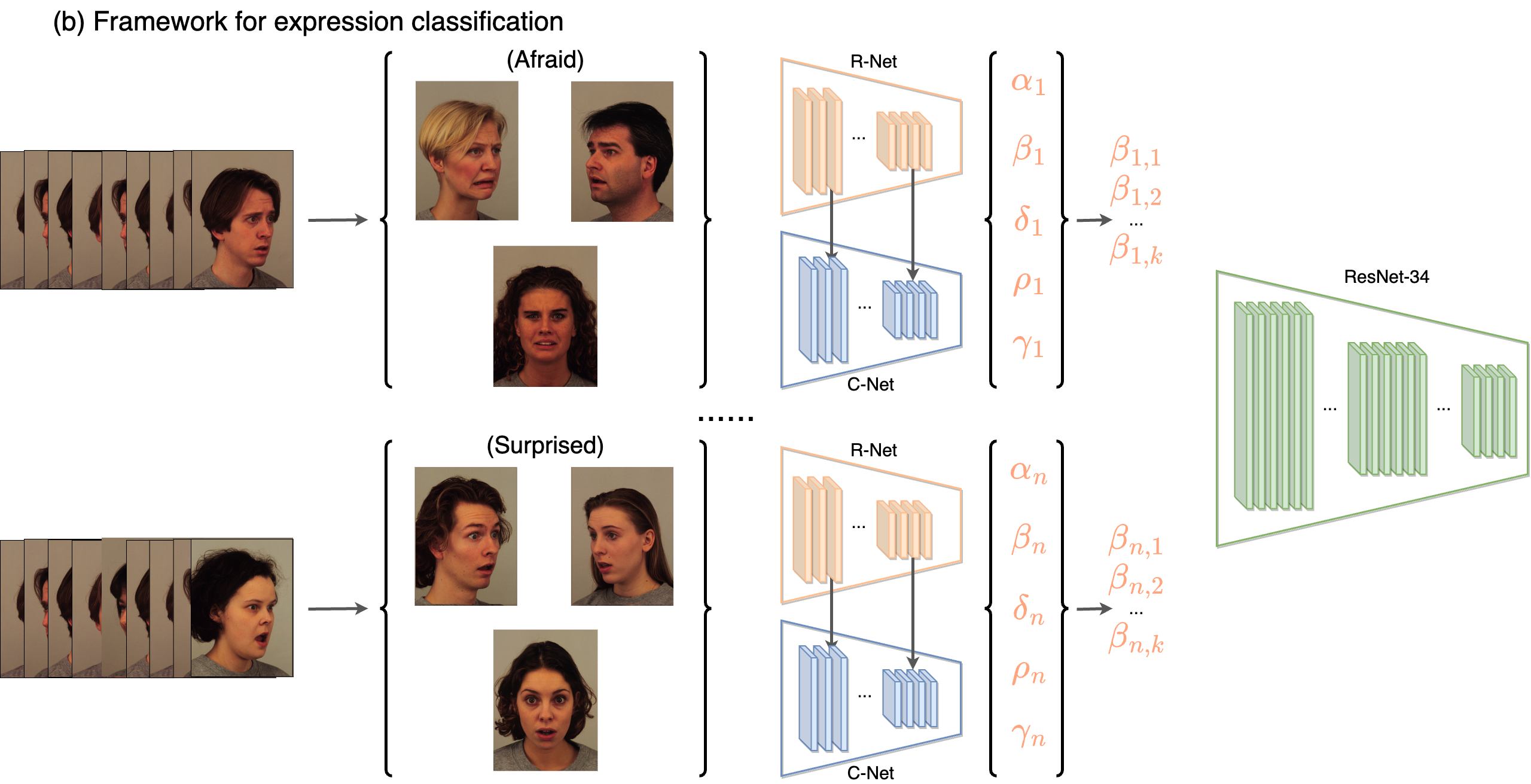}
  \caption{Framework for Expression Classification using Convolutional Neural Networks. The process is similar as previous figure.}
\end{figure}

\subsection{Overview of the 3D Morphable Model (3DMM)}

In this work, we employ the 3D Morphable Model (3DMM)~\cite{blanz1999morphable} as a foundational tool for 3D facial representation and reconstruction. This subsection provides a brief overview of 3DMM, which was originally proposed in \cite{blanz1999morphable} and has since become a widely used statistical model in facial analysis tasks.

The 3DMM represents facial geometry and texture using a compact parametric form derived from large-scale datasets of 3D facial scans. Specifically, the 3D facial shape $\textbf{S} \in \mathbb{R}^{3N}$ and texture $\textbf{T} \in \mathbb{R}^{3N}$ are modeled as linear combinations of a mean component and principal components obtained via Principal Component Analysis (PCA):
\begin{equation}
    \textbf{S} = \bar{\textbf{S}} + \textbf{B}_{id} \alpha + \textbf{B}_{exp} \beta,
    \label{eq:shape_model}
\end{equation}
\begin{equation}
    \textbf{T} = \bar{\textbf{T}} + \textbf{B}_t \delta,
    \label{eq:texture_model}
\end{equation}
where $\bar{\textbf{S}}$ and $\bar{\textbf{T}}$ represent the average shape and texture, $\textbf{B}_{id}$ and $\textbf{B}_{exp}$ are PCA bases for identity and expression variations, and $\textbf{B}_t$ is the PCA basis for texture. The coefficient vectors $\alpha$, $\beta$, and $\delta$ encode identity, expression, and texture, respectively, providing a compact representation of individual-specific facial features.

The identity and texture bases are derived from the Basel Face Model (BFM)~\cite{paysan20093d}, while the expression basis is obtained from \cite{guo2018cnn}. These bases are computed on datasets of 3D facial scans, capturing key modes of variation in facial shapes, expressions, and textures.

To account for variations in lighting, the 3DMM integrates the Spherical Harmonics (SH) model~\cite{ramamoorthi2001efficient,ramamoorthi2001signal}, which represents low-frequency illumination effects. The appearance of a vertex is computed as:
\begin{equation}
    C(n_i, t_i | \gamma) = t_i \cdot \sum_{b=1}^{B^2} \gamma_b \Phi_b(n_i),
    \label{eq:illumination}
\end{equation}
where $\Phi_b$ are the SH basis functions, and $\gamma_b$ are the SH coefficients that parameterize the lighting environment.

The 3DMM also incorporates a perspective camera model to project the 3D face onto a 2D image plane. The pose is parameterized by a rotation matrix $R \in \text{SO}(3)$ and a translation vector $t \in \mathbb{R}^3$, allowing the 2D projection of a vertex $s_i$ to be expressed as:
\begin{equation}
    \mathbf{p}_i = \text{Proj}(s_i | R, t) = \textbf{K} \cdot (R \cdot s_i + t),
    \label{eq:camera_model}
\end{equation}
where $\textbf{K}$ is the intrinsic camera matrix.

The fitting process involves estimating the parameters $\{\alpha, \beta, \delta, R, t, \gamma\}$ by minimizing the discrepancy between the observed 2D image and the image rendered by the 3DMM using the illumination and camera models. For more details on 3DMM, we refer the reader to \cite{blanz1999morphable}, \cite{paysan20093d}, and \cite{ramamoorthi2001efficient}.

\subsection{Overview of R-Net and C-Net for 3D Face Reconstruction}

This section summarizes the 3D face reconstruction approach proposed in \cite{deng2019accurate}, which employs R-Net and C-Net for hybrid- and weakly-supervised learning. We include this description for completeness.

\paragraph{R-Net for Single-Image Reconstruction.}  
R-Net is used to regress a coefficient vector $\mathbf{x}$ from a single RGB image $I$, enabling analytical generation of a reconstructed image $I'$ through differentiable mathematical operations. The network is trained using a combination of photometric, landmark, perceptual, and regularization losses.

The photometric loss ensures pixel-level consistency between $I$ and $I'$:

\[
L_{\text{photo}}(\mathbf{x}) = \frac{\sum_{i \in M} A_i \cdot \left\lVert I_i - I'_i(\mathbf{x}) \right\rVert_{2}}{\sum_{i \in M} A_i},
\]

where $M$ is the reprojected face region, and $A_i$ is a skin color-based attention mask computed using a naive Bayes classifier trained on skin-color data from \cite{jones2002statistical}. For robustness to occlusions and variations (e.g., makeup or beards), $A_i$ is defined as:

\[
A_i =
\begin{cases}
1, & \text{if } P_i > 0.5, \\
P_i, & \text{otherwise}.
\end{cases}
\]

R-Net also incorporates a landmark loss to align 3D landmarks with detected 2D landmarks $\{q_n\}$ from the image:

\[
L_{\text{lan}}(\mathbf{x}) = \frac{1}{N} \sum_{n=1}^{N} \omega_n \left\lVert q_n - q'_n(\mathbf{x}) \right\rVert^2,
\]

where $\{q'_n\}$ are the projected 3D landmarks, and $\omega_n$ are the landmark weights.

To capture global consistency, a perceptual loss is defined based on deep feature similarity:

\[
L_{\text{per}}(\mathbf{x}) = 1 - \frac{\langle f(I), f(I'(\mathbf{x})) \rangle}{\|f(I)\| \cdot \|f(I'(\mathbf{x}))\|},
\]

where $f(\cdot)$ represents FaceNet-based feature extraction \cite{schroff2015facenet}. Regularization on the 3D Morphable Model (3DMM) coefficients prevents degeneration:

\[
L_{\text{coef}}(\mathbf{x}) = \omega_\alpha \|\alpha\|^2 + \omega_\beta \|\beta\|^2 + \omega_\gamma \|\delta\|^2.
\]

Additionally, a texture flattening constraint is introduced to reduce variance in the texture map:

\[
L_{\text{tex}}(\mathbf{x}) = \sum_{c \in \{r,g,b\}} \text{var}(T_c, R(\mathbf{x})),
\]

where $R$ is a predefined skin region covering the cheeks, nose, and forehead. The overall loss combines these components to train R-Net to produce accurate 3D reconstructions from single images. For further details, please refer to \cite{deng2019accurate}.

\paragraph{C-Net for Multi-Image Reconstruction.}  
To aggregate 3D face information from multiple images of the same subject, the method employs a weakly-supervised neural aggregation network, C-Net. Given a set of images $\mathcal{I} := \{I^j | j = 1, \ldots, M\}$, C-Net predicts a confidence vector $\mathbf{c} \in \mathbb{R}^{80}$ for the identity-related shape coefficients $\alpha^j \in \mathbb{R}^{80}$ of each image $j$. The final aggregated shape coefficient $\alpha_{\text{aggr}}$ is computed as:

\[
\alpha_{\text{aggr}} = \left(\sum_{j} \mathbf{c}^j \odot \alpha^j\right) \oslash \left(\sum_{j} \mathbf{c}^j\right),
\]

where $\odot$ and $\oslash$ denote element-wise multiplication and division, respectively.

C-Net is trained without explicit labels by generating reconstructed images $\{I'^j\}$ from the aggregated coefficients. The training loss is defined as:

\[
L(\{\hat{\mathbf{x}}^j\}) = \frac{1}{M} \sum_{j=1}^{M} L(\hat{\mathbf{x}}^j),
\]

where $L(\cdot)$ is the hybrid-level loss function described above. The aggregation process is fully differentiable, allowing the backpropagation of error to the confidence vector $\mathbf{c}$ and the C-Net weights.

The design of C-Net and its training scheme are inspired by set-based face recognition methods \cite{yang2017neural}. By leveraging pose differences across the image set, C-Net improves shape reconstruction accuracy. To enhance efficiency, C-Net reuses feature maps from R-Net. For further details, please refer to \cite{deng2019accurate}.

\subsection{Facial classification models}
In this subsection, we describe our proposed facial classification models in details.

\subsubsection{Individual and Gender Classification Network}

In this work, we propose a classification framework that leverages ResNet-34, a deep convolutional neural network architecture, to perform individual and gender classification based on 3D facial shape parameters extracted using the 3D Morphable Model (3DMM). ResNet-34 is well-suited for this task due to its ability to learn complex hierarchical features and its proven success in various image classification challenges. Below, we describe the training and structure of the classification network, using gender classification as a detailed example. The same approach is extended to individual classification.

\paragraph{Data Preparation.}  
We assume that the training data consists of a collection of 2D facial images captured from multiple views for each individual. For gender classification, we organize the data into two categories: male and female. For each image, the facial features $\{\alpha, \beta, \delta, \gamma, p\}$ are extracted using the R-Net and aggregated into a unified set of facial parameters using the C-Net. These aggregated features represent a compact and robust description of the individual’s face, incorporating shape, expression, and pose information. The focus of this classification task is on the 80-dimensional shape coefficient vector $\alpha$, which encodes identity-related facial shape features.

\paragraph{Feature Extraction and Input Representation.}  
The shape coefficient vector $\alpha$, output by the C-Net, is used as the input to the ResNet-34 classification model. ResNet-34 is a residual network that employs shortcut connections to facilitate deep learning and alleviate the vanishing gradient problem, allowing the network to effectively learn from the data. The input $\alpha$ vectors are normalized and fed into the network, which processes them through a series of convolutional, activation, and pooling layers to extract high-level features. These features are then passed through fully connected layers to produce a final prediction for the classification task.

\paragraph{Model Architecture.}  
The ResNet-34 architecture consists of 34 layers, including convolutional layers, batch normalization layers, ReLU activation functions, and identity shortcut connections. The use of residual connections allows the network to learn residual mappings, which are easier to optimize compared to direct mappings. This design enables ResNet-34 to capture complex patterns in the input data while maintaining computational efficiency. For our task, the network is adapted to accept the 80-dimensional $\alpha$ vectors as input and output a prediction for the relevant classification category (e.g., male or female for gender classification).

\paragraph{Training Procedure.}  
The ResNet-34 model is trained using a labeled dataset of shape coefficient vectors $\alpha$ corresponding to male and female faces. The training process involves optimizing the model’s parameters to minimize the cross-entropy loss function, which is commonly used for multi-class classification problems. The cross-entropy loss is defined as:
\begin{equation}
    \mathcal{L}_{\text{CE}}(p, q) = -\sum_{i} p_i \log(q_i),
\end{equation}
where $p$ represents the true class probability distribution, and $q$ represents the predicted class probability distribution. This loss function encourages the model to maximize the likelihood of correctly predicting the true class for each input.

To further improve the model’s performance, we employ data augmentation techniques such as random rotations and scaling to increase the diversity of the training data, thereby reducing overfitting. Additionally, we use a learning rate scheduler to adjust the learning rate dynamically during training, ensuring stable convergence.

\paragraph{Classification Output.}  
The final output of the ResNet-34 network is a probability distribution over the possible classes (e.g., male and female). During inference, the class with the highest predicted probability is selected as the network’s classification result. The use of the cross-entropy loss function ensures that the network learns to assign higher probabilities to the correct classes while minimizing the likelihood of incorrect predictions.

\paragraph{Extension to Individual Classification.}  
The same framework can be extended to perform individual classification, where the task is to distinguish between different individuals based on their 3D facial shape features. In this case, the training data consists of labeled shape coefficient vectors $\alpha$ for each individual in the dataset. The ResNet-34 network is trained to classify these vectors into the corresponding individual identity classes. The number of output classes in the final layer of the network is adjusted to match the number of individuals in the dataset.

By leveraging the robust feature extraction capabilities of the 3DMM and the powerful classification capabilities of ResNet-34, our framework achieves high accuracy across both individual and gender classification tasks. The combination of multi-view data, aggregated features, and deep-learning-based classification allows for a reliable and effective solution to these facial analysis challenges.

An overview of the framework is illustrated in Figure \ref{individualgender}.

\subsubsection{Expression Classification Network}

For the task of expression classification, we propose a framework that utilizes 3D facial parameters obtained from R-Net and C-Net in conjunction with a deep neural network to classify human faces into various expression categories. The main goal is to leverage the robust feature representation provided by the 3D Morphable Model (3DMM) and implement a ResNet-32 architecture to identify and classify facial expressions effectively. Below, we describe the key components of this framework in detail.

\paragraph{Data Preparation.}  
To train the expression classification network, we use a dataset containing human face images with diverse expressions such as happy, sad, surprised, angry, and neutral. For each subject in the dataset, multiple 2D images are captured from different viewpoints, ensuring that the data incorporates variations in pose and perspective. Using the R-Net, we extract the facial parameters $\{\alpha, \beta, \delta, \gamma, p\}$ from each image, where $\alpha$ represents the shape, $\beta$ captures the expression, $\delta$ models the texture, $\gamma$ encodes illumination, and $p$ accounts for pose. These parameters are then aggregated using C-Net to produce a unified representation of each face, emphasizing consistency across different views.

For expression classification, the focus is on the expression coefficient vector $\beta$, which captures the primary variations in facial expression. By isolating $\beta$ from the other parameters, we ensure that the features input to the network are specifically tailored to the expression classification task.

\paragraph{Network Architecture.}  
The expression classification network is built on the ResNet-32 architecture, a lightweight convolutional neural network designed to learn discriminative features efficiently. ResNet-32 utilizes residual connections to enable effective training of deep networks by mitigating the vanishing gradient problem. This property is particularly important for learning subtle variations in facial expressions. The input to the network is the 64-dimensional expression coefficient vector $\beta$. The network processes this input through a series of convolutional layers, residual blocks, and fully connected layers to extract high-level features that are critical for distinguishing between different expressions.

The final layer of the network consists of a softmax activation function that outputs a probability distribution over the predefined expression classes. The class with the highest probability is selected as the predicted expression for the given input.

\paragraph{Training Procedure.}  
The ResNet-32 model is trained using labeled expression data, where each $\beta$ vector is annotated with its corresponding expression class. The training process involves minimizing the cross-entropy loss function:
\begin{equation}
    \mathcal{L}_{\text{CE}}(p, q) = -\sum_{i} p_i \log(q_i),
\end{equation}
where $p$ represents the true class probability distribution, and $q$ is the predicted class probability distribution. This loss function guides the network to assign higher probabilities to the correct expression classes while minimizing the likelihood of incorrect predictions.

To enhance the model’s ability to generalize, we apply data augmentation by simulating variations in lighting conditions, pose, and image quality. This ensures that the network learns robust features that are invariant to these factors. Additionally, a learning rate scheduler is used to adjust the learning rate dynamically during training, promoting stable convergence and preventing overfitting.

\paragraph{Feature Learning for Expression Classification.}  
One of the key advantages of using ResNet-32 in this framework is its ability to identify the most important features from the input $\beta$ vectors for expression classification. By processing the 64-dimensional expression coefficients through multiple layers, the network learns hierarchical features that capture subtle differences between expressions. These features are then used by the fully connected layers to classify the input into the appropriate expression category.

\paragraph{Classification Output.}  
During inference, the trained ResNet-32 model takes the expression coefficient vector $\beta$ as input and outputs a probability distribution over the expression classes. The class with the highest probability is selected as the predicted expression. This approach ensures that the classification is based on the most relevant features for the given task, resulting in high accuracy and reliability.

\paragraph{Framework Summary.}  
The expression classification network integrates 3D facial parameters extracted using R-Net and C-Net with ResNet-32 to achieve robust and accurate expression recognition. By focusing on the expression coefficient $\beta$ and leveraging the feature extraction capabilities of ResNet-32, the framework effectively distinguishes between various facial expressions. This approach demonstrates the potential of combining 3DMM-based facial features with deep learning for challenging facial analysis tasks. An overview of the framework is illustrated in Figure \ref{expression}.


\section{Experimental Datasets and Setup}
\label{sec:experiments}
We first provide an overview of the datasets used in our experiments and the setup employed for evaluating the performance of our proposed method in various facial classification tasks. Specifically, we utilized two publicly available datasets, the Karolinska Directed Emotional Faces (KDEF) dataset and the FEI Face Database, to assess the effectiveness of using 3DMM coefficients ($\alpha$ and $\beta$) for face recognition, expression classification, and gender classification tasks.

\paragraph{KDEF Dataset.}  
The Karolinska Directed Emotional Faces (KDEF) dataset~\cite{lundqvist1998karolinska}, developed by Lundqvist, Flykt, and Öhman in 1998, is a comprehensive collection consisting of 4900 high-quality images of 70 individuals (35 females and 35 males) aged between 20 and 30 years. Each individual in the dataset displays seven distinct emotional expressions: afraid, angry, disgust, happy, neutral, sad, and surprised. These expressions are captured from five angles (-90°, -45°, 0°, +45°, +90°), providing a rich multi-view dataset that is well-suited for facial classification studies.

The individuals featured in the dataset were carefully chosen based on specific criteria, such as the absence of facial hair, eyeglasses, and noticeable makeup, to ensure a consistent and neutral appearance. Before the images were captured, participants rehearsed their expressions to ensure that the emotional displays were both natural and clearly discernible. The photographic process employed a Pentax LX camera equipped with a 135 mm lens, and Kodak 320 T film was used to achieve high-quality results. The lighting setup was meticulously designed to provide uniform illumination across the participant's face, minimizing the presence of harsh shadows or uneven lighting.

The digitization process involved the use of a Macintosh 8500/120 computer and a Polaroid SprintScan 35 scanner. Images were scanned in RGB color at a resolution of 625 dpi and post-processed using Adobe Photoshop 4 to enhance their quality. The final images were cropped to a resolution of 562 × 762 pixels and stored in JPEG format. Each file was carefully coded to include information about the participant's session, gender, identity, expression, and viewing angle, enabling easy identification and organization of the dataset.

The KDEF dataset provides a robust testbed for evaluating facial classification models, particularly in tasks such as face recognition and expression classification, due to its diverse range of expressions, multi-view images, and controlled environmental conditions.

\begin{figure}[H]
    \centering
    \begin{minipage}{0.18\textwidth}
        \centering
        \includegraphics[width=\linewidth]{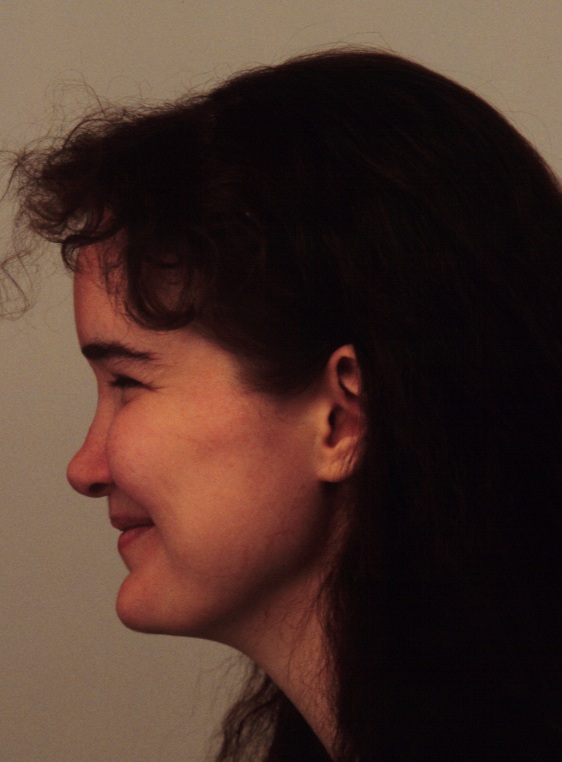}
    \end{minipage}\hfill
    \begin{minipage}{0.18\textwidth}
        \centering
        \includegraphics[width=\linewidth]{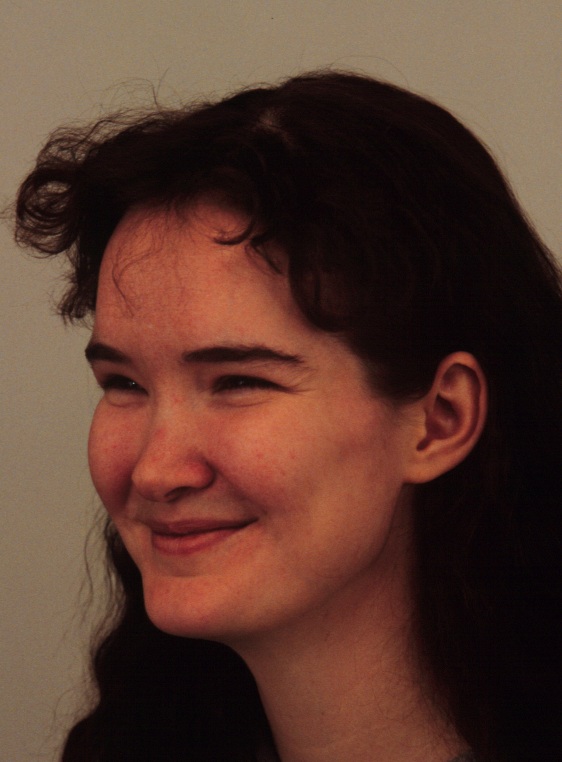}
    \end{minipage}\hfill
    \begin{minipage}{0.18\textwidth}
        \centering
        \includegraphics[width=\linewidth]{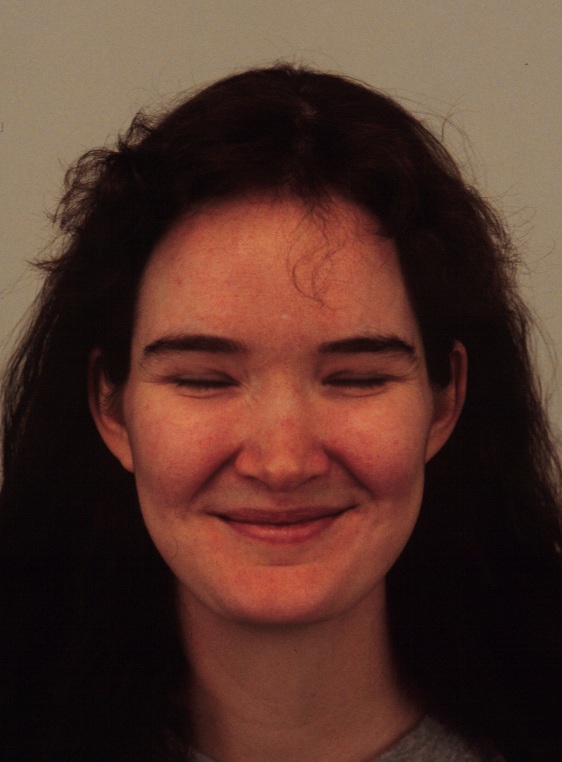}
    \end{minipage}\hfill
    \begin{minipage}{0.18\textwidth}
        \centering
        \includegraphics[width=\linewidth]{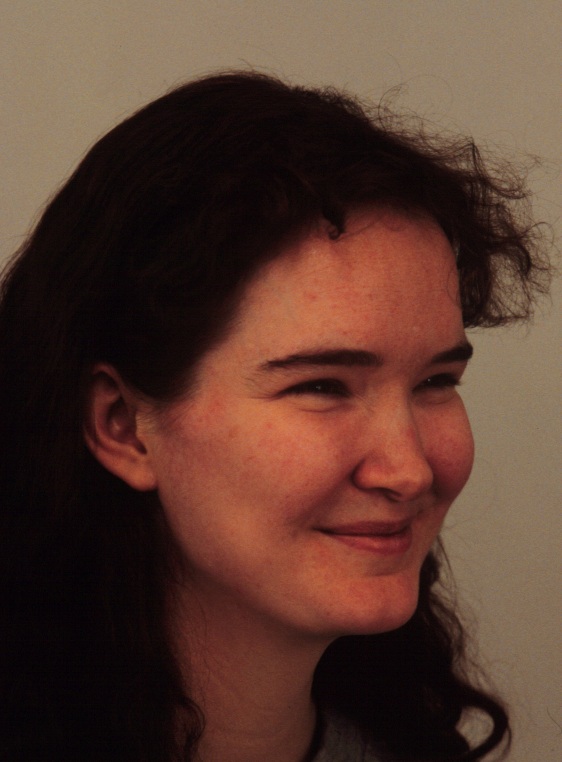}
    \end{minipage}\hfill
    \begin{minipage}{0.18\textwidth}
        \centering
        \includegraphics[width=\linewidth]{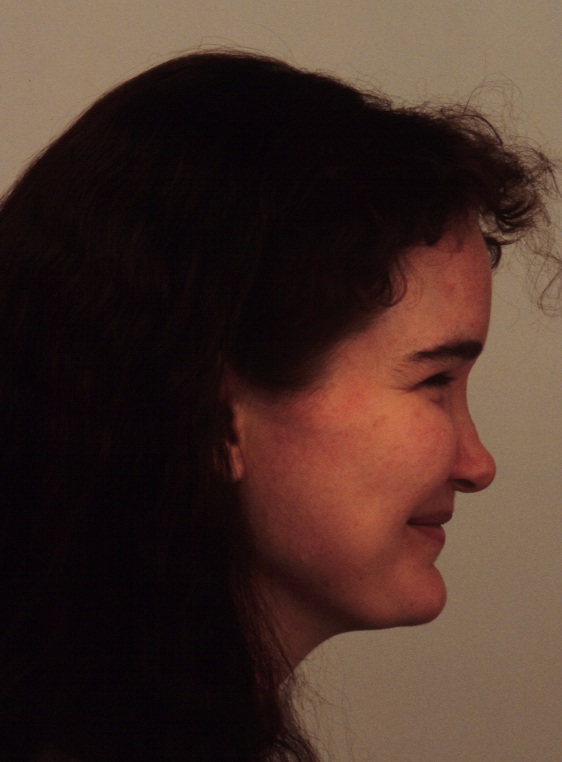}
    \end{minipage}
    
    \centering
    \begin{minipage}{0.18\textwidth}
        \centering
        \includegraphics[width=\linewidth]{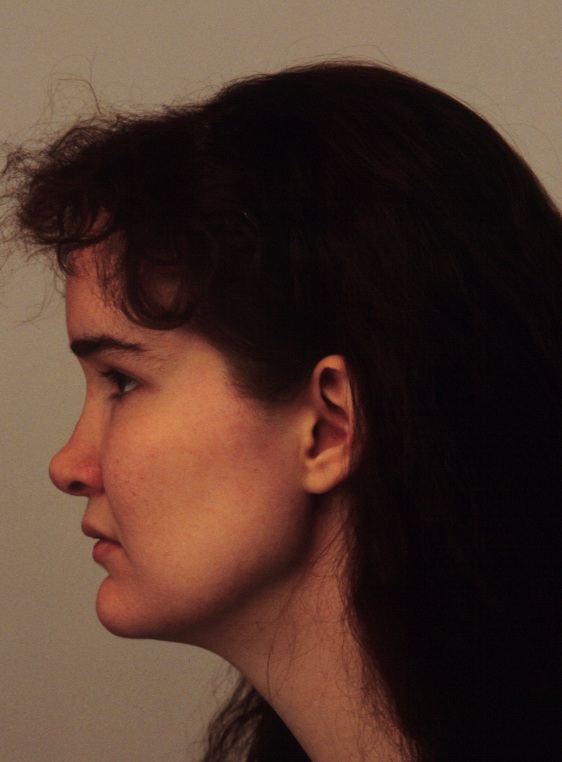}
    \end{minipage}\hfill
    \begin{minipage}{0.18\textwidth}
        \centering
        \includegraphics[width=\linewidth]{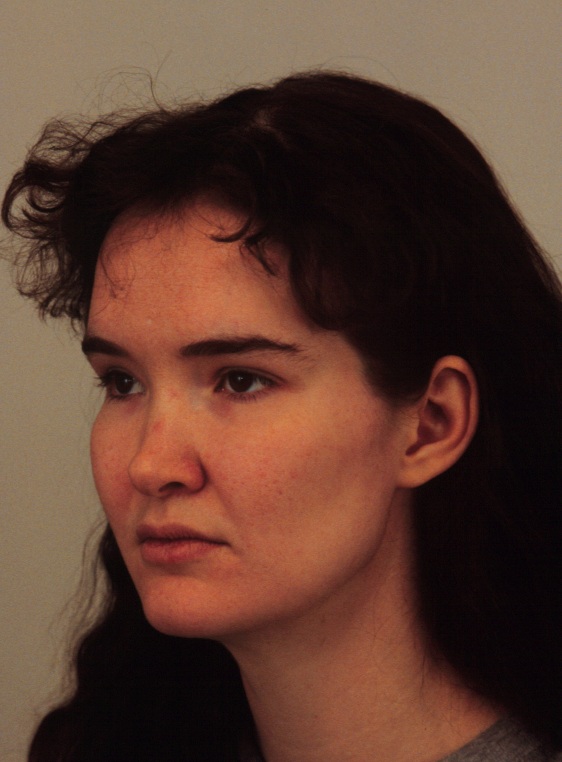}
    \end{minipage}\hfill
    \begin{minipage}{0.18\textwidth}
        \centering
        \includegraphics[width=\linewidth]{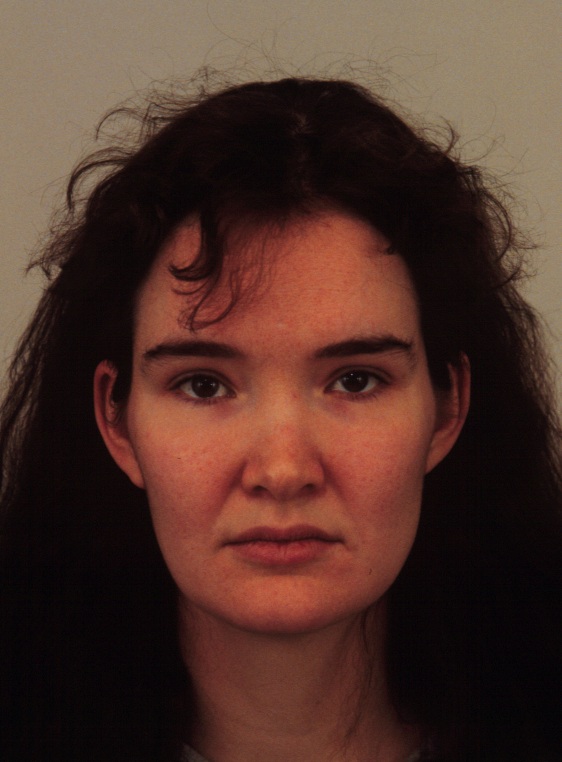}
    \end{minipage}\hfill
    \begin{minipage}{0.18\textwidth}
        \centering
        \includegraphics[width=\linewidth]{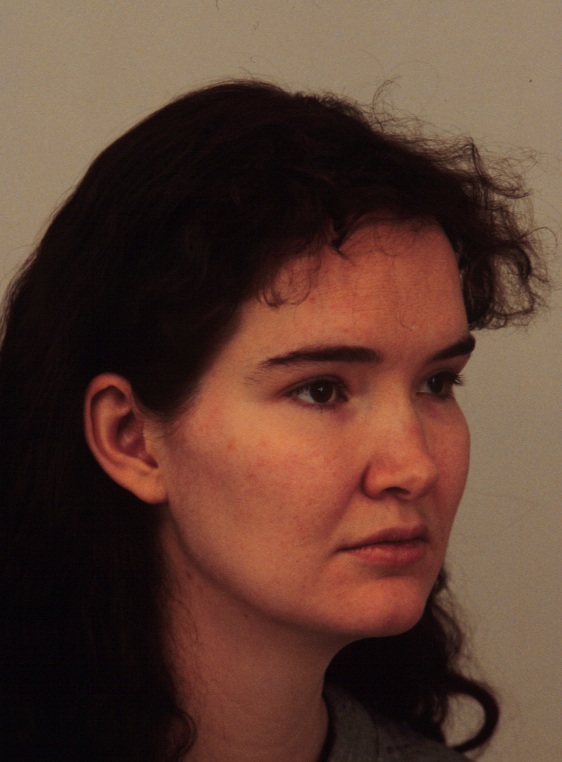}
    \end{minipage}\hfill
    \begin{minipage}{0.18\textwidth}
        \centering
        \includegraphics[width=\linewidth]{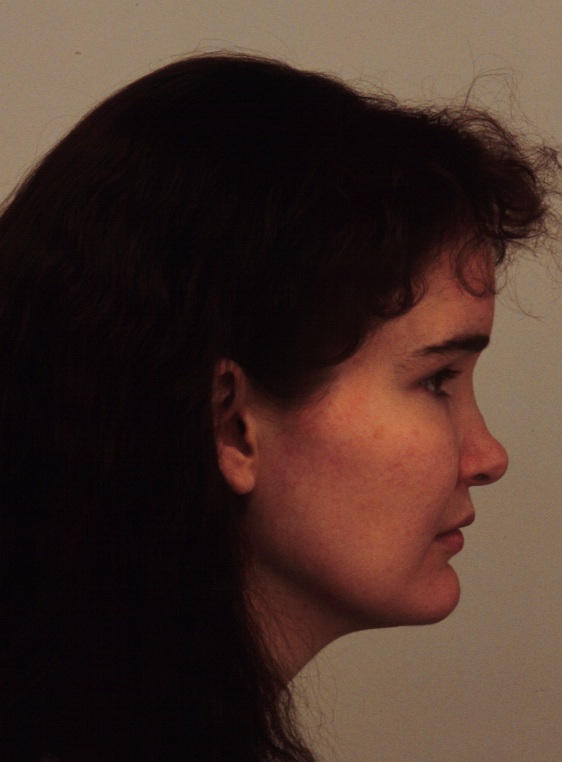}
    \end{minipage}
    
    \vspace{0.5cm}
    \centering
    \caption{Examples from KDEF Dataset (Samples: F04HA and F04NE)}
\end{figure}

\paragraph{FEI Face Database.}  
The FEI Face Database~\cite{FEI2017FaceDB} is a Brazilian facial image dataset developed at the Artificial Intelligence Laboratory of FEI in São Bernardo do Campo, São Paulo, Brazil. The dataset was collected between June 2005 and March 2006 and features 2800 colorful images of 200 individuals, evenly distributed between 100 males and 100 females. For each individual, 14 images were captured in an upright frontal position with varying degrees of profile rotation, ranging up to approximately 180°.

The images were taken against a uniform white background, ensuring consistent lighting conditions and minimal distractions. While the scale of the faces may vary slightly (within a 10\% range), the uniform environmental conditions make the dataset suitable for controlled experiments. Each image is stored at a resolution of 640 × 480 pixels.

The participants in the database include students and staff members of FEI, aged between 19 and 40 years, exhibiting a diversity of appearances, hairstyles, and accessories. This diversity makes the FEI Face Database an excellent resource for studying gender classification and other facial recognition tasks. The equal representation of male and female participants ensures a balanced dataset for evaluating gender classification models.

\begin{figure}[H]
    \centering
    \begin{minipage}{0.22\textwidth}
        \centering
        \includegraphics[width=\linewidth]{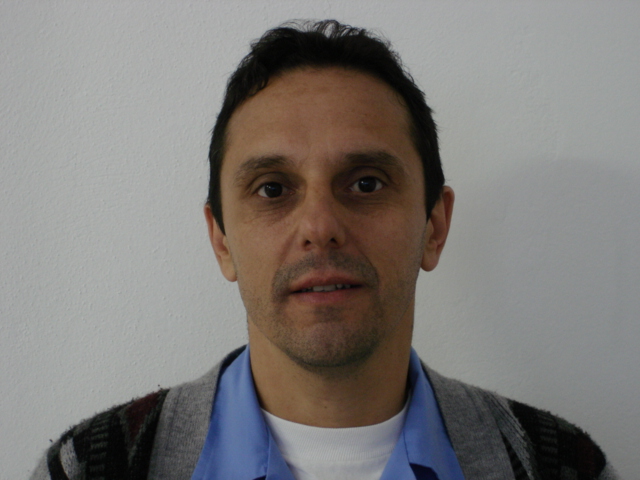}
    \end{minipage}\hfill
    \begin{minipage}{0.22\textwidth}
        \centering
        \includegraphics[width=\linewidth]{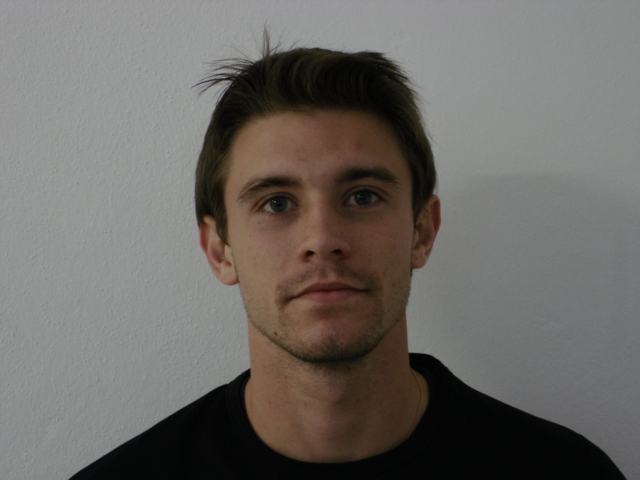}
    \end{minipage}\hfill
    \begin{minipage}{0.22\textwidth}
        \centering
        \includegraphics[width=\linewidth]{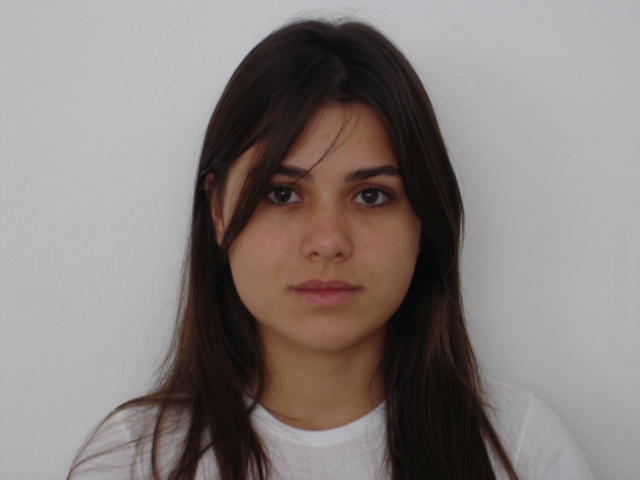}
    \end{minipage}\hfill
    \begin{minipage}{0.22\textwidth}
        \centering
        \includegraphics[width=\linewidth]{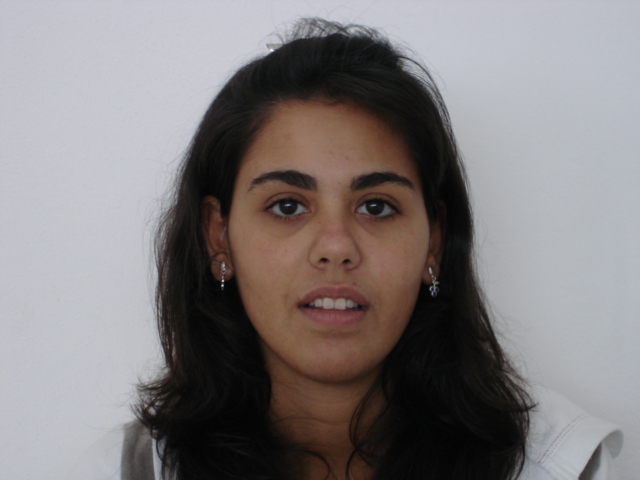}
    \end{minipage}\hfill
    
    \vspace{0.5cm}
    \centering
    \caption{Examples from FEI Face Database}
\end{figure}

\paragraph{Experimental Setup.}  
For our experiments, we used the $\alpha$ and $\beta$ coefficients derived from images in both datasets as inputs to our proposed classification framework. The coefficients were extracted using R-Net, and multi-view information was aggregated using C-Net to create unified representations of each subject’s face. The experiments were conducted on a MacBook Air equipped with an Apple M1 chip, with Pytorch leveraging Metal Performance Shaders (MPS) for GPU-accelerated training. The setup allowed us to evaluate both the accuracy and computational efficiency of our approach across different facial analysis tasks.

\subsection{Face Recognition (KDEF)}

For this experiment, we used the KDEF dataset, which consists of 4900 images of facial expressions from 70 individuals, each displaying 7 distinct emotional expressions captured from 5 different angles. This dataset is particularly suitable for evaluating face recognition models due to its diverse range of expressions and multi-view images for each subject.

To simplify the experiment, we selected 10 individuals from the dataset and divided their images into a training set and a validation set. Specifically, the training set consisted of 200 images (20 images per individual), and the validation set consisted of 150 images (15 images per individual). For face recognition, we investigated two distinct input representations: the original images and the $\alpha$ coefficients derived from the 3DMM.

To generate $\alpha$ coefficients, we used R-Net to extract the individual shape coefficients from each image and utilized C-Net to aggregate these coefficients. For each individual, we generated $\alpha$ coefficients from the 20 training images, then selected 3 $\alpha$ coefficients at a time to create combined representations. This process was repeated $\binom{20}{3} = 1140$ times, resulting in 1140 combined $\alpha$ coefficients for training. Similarly, we generated $\binom{15}{3} = 455$ combined $\alpha$ coefficients for validation from the 15 validation images per individual.

The results of this experiment revealed that $\alpha$ coefficients significantly outperformed the original images in terms of classification accuracy. Specifically, the classification using original images achieved an accuracy of 87.3\%, while the classification using $\alpha$ coefficients achieved a perfect accuracy of 100\%. We also performed cross-validation to confirm the reliability of these results, and the accuracy remained consistent across all folds.

In addition to accuracy, computational efficiency was another key metric of interest. The average training time per input was approximately 0.054 seconds when using the original images, compared to only 0.005 seconds per input when using $\alpha$ coefficients. This significant reduction in training time highlights the computational advantage of using 3DMM-derived features, which offer a compact yet highly discriminative representation of facial shapes.

\subsection{Classification of Expressions (KDEF)}

The second experiment focused on classifying different emotional expressions using the KDEF dataset. This task involves distinguishing between 7 emotional categories: happy, sad, surprised, angry, disgusted, afraid, and neutral. The diversity of expressions in the dataset, combined with the multi-view images for each individual, makes it an excellent benchmark for evaluating expression classification models.

To prepare the data, we organized the images into 7 categories, each corresponding to one of the emotional expressions. The dataset was then split into a training set containing 2975 images and a validation set containing 1925 images. We conducted experiments using two input representations: the original images and the $\beta$ coefficients (expression coefficients) extracted from the 3DMM.

For each image, the $\beta$ coefficients were computed using R-Net and aggregated across multiple views using C-Net to produce a unified representation. These coefficients were then input into the expression classification network based on ResNet-32, which was trained to classify the input into one of the 7 expression categories.

The experimental results demonstrated a clear advantage of using $\beta$ coefficients over the original images. Specifically, the classification using $\beta$ coefficients achieved an average accuracy of 83.5\% after cross-validation, whereas the classification using original images achieved a lower accuracy of 70.6\%. This improvement can be attributed to the ability of $\beta$ coefficients to capture subtle variations in facial expressions more effectively than raw pixel values.

We also examined the computational efficiency of the two approaches. The training time per epoch ranged from 150.76 to 219.61 seconds when using the original images. In contrast, the training time was significantly reduced to a range of 4.05 to 4.33 seconds per epoch when using $\beta$ coefficients. This drastic reduction in training time further emphasizes the computational benefits of leveraging 3DMM-derived features for expression classification.

\subsection{Classification of Gender (FEI Face Database)}

The final experiment focused on gender classification using the FEI face database. This dataset contains 2800 colorful face images of 200 individuals, evenly distributed between 100 males and 100 females. All images were captured in an upright frontal position with profile rotation of up to about 180 degrees, providing sufficient variation for evaluating gender classification models.

We divided the dataset into a training set and a validation set. The training set consisted of 50 male and 50 female images, while the validation set contained the remaining 50 male and 50 female images. As with the other experiments, we conducted separate trials using the original images and the $\alpha$ coefficients.

For each image, $\alpha$ coefficients were extracted using R-Net and aggregated using C-Net to create a unified representation. These coefficients were then input into the gender classification network based on ResNet-34, which was trained to classify the input into the male or female category.

The results showed a significant improvement in classification accuracy when using $\alpha$ coefficients. Specifically, the classification using original images achieved an accuracy of 79.8\%, while the classification using $\alpha$ coefficients achieved an average accuracy of 95.4\% after cross-validation. This improvement highlights the effectiveness of shape-based features for capturing gender-specific facial characteristics.

In terms of computational efficiency, the training time per epoch ranged from 79.10 to 99.56 seconds when using the original images. However, the training time was reduced to a range of 38.18 to 44.56 seconds per epoch when using $\alpha$ coefficients. This reduction in training time further demonstrates the practicality of using 3DMM features for gender classification.

\section{Analysis}

The experimental results demonstrate that using $\alpha$ and $\beta$ coefficients as inputs for classification tasks yields superior accuracy and significantly reduces training time compared to using the original images directly. This highlights the critical role of 3DMM-based shape and expression representations in capturing the most discriminative facial features for various classification tasks.

The following subsections provide a detailed analysis of the results, including comparative accuracy, training time efficiency, expression-specific performance, and future directions for improvement.

\subsection{Classification Results}

\begin{table}[ht]
\centering
\begin{tabular}{lcc}
\toprule
\textbf{Classification Task} & \textbf{Using Images} & \textbf{Using 3DMM Coefficients} \\
\midrule
Face Recognition (KDEF) & 87.3\% & 100\% \\
Classify Expression (KDEF) & 70.6\% & 83.5\% \\
Classify Gender (FEI) & 79.8\% & 95.4\% \\
\bottomrule
\end{tabular}
\caption{Classification Results Using Different Inputs}
\label{tab:classification_results}
\end{table}

Table~\ref{tab:classification_results} summarizes the classification accuracy achieved using two different input representations: the original images and the $\alpha$/$\beta$ coefficients derived from the 3DMM. Across all tasks, the use of 3DMM coefficients significantly outperformed the use of raw images. For face recognition, the accuracy improved from 87.3\% to a perfect 100\%. Similarly, expression classification accuracy increased from 70.6\% to 83.5\%, while gender classification saw a substantial improvement from 79.8\% to 95.4\%. The results underscore the effectiveness of the 3DMM-based representations in capturing essential facial features and discriminative information for classification tasks.

The superior performance of the 3DMM coefficients can be attributed to their ability to disentangle identity, expression, and other facial variations, enabling the models to focus on the most relevant features for each classification task. This is especially important when dealing with multi-view datasets, as the aggregated coefficients provide a unified and robust representation of the 3D facial structure.

\subsection{Training Time Comparison}

\begin{table}[ht]
\centering
\begin{tabular}{lcc}
\toprule
\textbf{Classification Task} & \textbf{Using Images} & \textbf{Using 3DMM Coefficients} \\
\midrule
Face Recognition & 0.054s & 0.005s \\
Classify Expression & 0.062s & 0.001s \\
Classify Gender  & 0.006s & 0.003s \\
\bottomrule
\end{tabular}
\caption{Average Training Time (Per Input) Using Different Inputs}
\label{tab:training_time_comparison}
\end{table}

In addition to improving accuracy, the use of $\alpha$ and $\beta$ coefficients also leads to a significant reduction in training time. Table~\ref{tab:training_time_comparison} compares the average training time per input for both representations. For all three tasks, the training time using 3DMM coefficients is drastically lower than when using images. 

For instance, in face recognition, the average training time per input is reduced from 0.054 seconds (using images) to 0.005 seconds (using $\alpha$ coefficients), reflecting a 10-fold improvement in computational efficiency. Similarly, in expression classification, the training time is reduced from 0.062 seconds to just 0.001 seconds per input. This reduction in computational cost is particularly valuable for large-scale applications and real-time systems, where efficiency is critical.

The reduced training time can be attributed to the compact nature of the 3DMM coefficients, which significantly compress the input data while retaining the most important information. By reducing the dimensionality of the input features, the model can converge faster and require fewer resources during training.

\subsection{Expression Analysis}

\begin{table}[ht]
\centering
\begin{tabular}{lccccccc}
\toprule
& \textbf{AF} & \textbf{AN} & \textbf{DI} & \textbf{HA} & \textbf{NE} & \textbf{SA} & \textbf{SU}\\
\midrule
\textbf{AF} & \textbf{75.26\%} & 1.10\% & 3.87\% & 1.66\% & 2.40\% & 5.00\% & 10.70\% \\
\textbf{AN} & 4.24\% & \textbf{65.00\%} & 13.09\% & 1.11\% & 8.29\% & 7.56\% & 0.73\% \\
\textbf{DI} & 3.49\% & 2.76\% & \textbf{86.75\%} & 1.11\% & 0.18\% & 5.34\% & 0.37\% \\
\textbf{HA} & 0.74\% & 0.74\% & 1.29\% & \textbf{95.77\%} & 0.37\% & 0.74\% & 0.37\% \\
\textbf{NE} & 0.18\% & 1.83\% & 0\% & 0\% & \textbf{93.22\%} & 4.76\% & 0\% \\
\textbf{SA} & 3.12\% & 1.65\% & 4.58\% & 0.18\% & 9.71\% & \textbf{80.57\%} & 0.18\% \\
\textbf{SU} & 12.19\% & 0.55\% & 0\% & 0.19\% & 0.74\% & 0.55\% & \textbf{85.78\%} \\
\bottomrule
\end{tabular}
\caption{The Confusion Matrix Using Proposed Method in Expression Classification}
\label{tab:confusion_matrix}
\end{table}

The confusion matrix in Table~\ref{tab:confusion_matrix} provides a detailed breakdown of the performance for each emotional expression. The abbreviations correspond to the following emotions: AF (afraid), AN (angry), DI (disgust), HA (happy), NE (neutral), SA (sad), and SU (surprised). 

While the overall accuracy of expression classification is high, certain expressions, such as angry (AN) and afraid (AF), exhibit lower classification rates. For example, the angry class achieves an accuracy of 65.00\%, which is lower than the accuracy for other classes like happy (95.77\%) and disgust (86.75\%). This discrepancy may be due to the subtle and overlapping features of certain expressions, making them more challenging to distinguish. Misclassification may also arise from variations in pose, lighting, or inconsistencies in the dataset annotations.

\subsection{Future Work}

While the experiments confirm the effectiveness and efficiency of using 3DMM coefficients for facial classification tasks, there are several potential areas for future exploration. One promising direction is to extend this framework to handle more complex tasks, such as age estimation, emotion intensity analysis, or multi-label classification of simultaneous facial attributes. Additionally, incorporating advanced neural network architectures, such as transformers or attention-based models, could further enhance the performance of the proposed method.

Another area of interest is evaluating the robustness of the model under challenging conditions, such as occlusions, extreme lighting, or low-resolution images. By addressing these challenges, the framework could be adapted for real-world applications, such as video-based analysis or facial recognition in unconstrained environments.

Finally, future research could explore the integration of other 3D facial parameters, such as texture coefficients ($\delta$), into the classification pipeline. This would provide a more comprehensive understanding of how different aspects of the 3D facial model contribute to classification performance.

\section{Conclusion}
\label{sec:conclusion}
Our novel approach for facial analysis using shape-based representations and weakly-supervised neural aggregation achieved remarkable results in various tasks. By employing 3DMM for facial shape representation, we demonstrated the effectiveness of alpha and beta coefficients, achieving 100\% accuracy in individual classification, 95.4\% accuracy in gender classification, and 83.5\% in expression classification.

\bibliographystyle{unsrt}  


\clearpage
\newpage
\bibliography{references}

\end{document}